\documentclass[pmlr]{jmlr}

\RequirePackage{graphicx}
 \usepackage{booktabs}
\usepackage[margin=false, inline=true]{fixme}
\usepackage{wrapfig}

\renewcommand{\cite}{\citep}

\jmlrproceedings{PMLR}{Preprint}

\title[Evaluation of Survival Prediction for Deceased Donor Kidney Transplants]{Paired Recipient-based Evaluation of Survival Prediction for Deceased Donor Kidney Transplants}

\author{\Name{Misaki Matsuura}
       \Email{mxm1407@case.edu}\\ 
       \addr Department of Computer and Data Sciences\\
       Case Western Reserve University\\
       Cleveland, OH 44106, USA
       \AND
       \Name{Mohammadreza Nemati}
       \Email{mohammadreza.nemati@case.edu}\\ 
       \addr Department of Computer and Data Sciences\\
       Case Western Reserve University\\
       Cleveland, OH 44106, USA
       \AND
       \Name{Dulat Bekbolsynov}
       \Email{dulat.bekbolsynov@utoledo.edu}\\ 
       \addr Department of Medical Microbiology and Immunology\\
       University of Toledo\\
       Toledo, OH 43614, USA
       \AND
       \Name{Stanislaw Stepkowski}
       \Email{stanislaw.stepkowski@utoledo.edu}\\ 
       \addr Department of Medical Microbiology and Immunology\\
       University of Toledo\\
       Toledo, OH 43614, USA
       \AND
       \Name{Kevin S. Xu}
       \Email{ksx2@case.edu}\\ 
       \addr Department of Computer and Data Sciences\\
       Case Western Reserve University\\
       Cleveland, OH 44106, USA%
}

\begin{document}

\maketitle

\begin{abstract}
There has been significant interest in using machine learning algorithms to predict kidney transplant outcomes, such as the number of years until a graft inevitably fails. 
These prediction algorithms could possibly be used for pre-transplant donor-recipient matching to identify more compatible donors and recipients and thus improve post-transplant outcomes. 
In this study, we explore the use of survival prediction models trained on deceased donor kidney transplant data from the Scientific Registry of Transplant Recipients (SRTR).
We propose a novel paired recipient-based evaluation framework that compares graft outcomes between two recipients who received kidneys from the same deceased donor, allowing us to evaluate the \emph{counterfactual benefit} of changing the recipient for a certain donor.
We find that five different survival prediction models, ranging in complexity from linear to deep learning-based models, all result in
$\sim$60\% paired recipient-based accuracy. 
We further translate this accuracy into an interpretable quantity of post-transplant years gained.
We also highlight major limitations of the commonly used concordance index (C-index) metric for evaluating survival prediction accuracy in this setting
and demonstrate that our proposed paired recipient-based accuracy metric is more clinically relevant and better reflects real-world allocation settings. 
\end{abstract}

\section{Introduction}
Kidney transplantation is a life-saving intervention for patients with end-stage renal disease (ESRD), yet the demand for donor kidneys consistently exceeds supply. This persistent imbalance has made optimal kidney allocation a critical challenge in clinical practice and public policy. The current 250-nautical mile (NM) circle allocation system is based on blood group compatibility and the order of appearance of a candidate (potential recipient) on the national waiting list \cite{adler2021greater}. 
The kidney donor profile index (KDPI) is used to evaluate donors \cite{zens2018impact, kadatz2023benefits}, and the estimated post-transplant survival (EPTS) is applied to evaluate recipients \cite{clayton2014external}.
To address the lack of supply of donor kidneys, more sophisticated algorithms are needed to match donors and recipients to significantly improve kidney transplant outcomes. 
In response, researchers have increasingly turned to machine learning (ML)-based survival prediction models to inform more effective allocation strategies \cite{nemati2023predicting, Ali2025AITransplant, Na2025KDPIEPTS}.

Survival prediction models estimate the likelihood of graft failure over time and offer the potential to match kidneys with recipients who are likely to benefit the most, thereby improving overall outcomes and making better use of scarce donor kidneys. For example, the artificial intelligence (AI)-based United Kingdom (UK) Live-Donor Kidney Transplant Outcome Prediction was proposed for the best live-donor selection \cite{Ali2025AITransplant}. Machine learning (ML) models also effectively identified dropout risk at referral, evaluation, and waitlisting, thereby identifying high risk patients \cite{AlAwadhi2025MLDropout}. While traditional models such as the Cox proportional hazards have been widely used to predict outcomes \cite{wolfe2008lyft, ashby2017kidney}, the rise of AI and ML methods has opened new possibilities to capture complex interactions and improve predictive accuracy.

Despite progress in AI and ML technology, their application in predictive models of kidney allocation remains limited. Many studies prioritize statistical metrics with the concordance index (C-index) \cite{harrell1982evaluating} without adequately addressing generalizability or real-world utility. Moreover, predictive models are often evaluated in isolation without considering their downstream effects on allocation decisions. This disconnect makes it difficult to assess the true value of a model from a policy perspective.

\begin{wrapfigure}[16]{r}{0.34\textwidth}
  \vspace{-16pt}
  \centering
  \includegraphics[width=0.34\textwidth]{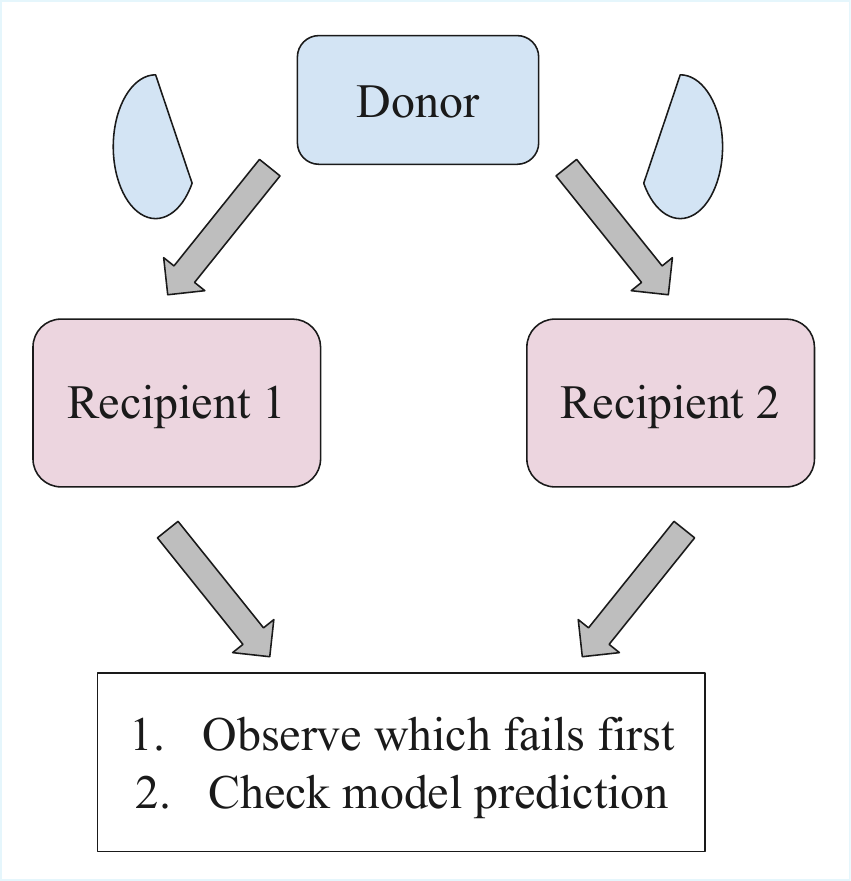}
  \caption{Our proposed paired recipient-based evaluation method.}
  \label{fig:eval_method}    
\end{wrapfigure}

Our study aims to bridge this gap by proposing a survival prediction framework that is both clinically interpretable and practically actionable in the context of kidney transplantation.
We make four main contributions in this paper:
\begin{enumerate}
\item We propose a new paired recipient-based evaluation metric for deceased donor kidney transplants: how frequently a survival prediction algorithm correctly predicts which of the two recipients from the same donor will have a longer-lasting graft, as shown in Figure~\ref{fig:eval_method}.

\item We provide a broad comparison of prediction accuracy across different survival prediction algorithms using both our paired recipient metric and the conventional evaluation metric for survival prediction, the C-index.

\item We translate paired recipient accuracy into prediction intervals on the post-transplant years gained by a particular survival prediction algorithm if used to select between the two transplant recipients.

\item We measure fairness of existing prediction algorithms, if used to select the recipient, from the perspective of the recipient's race.
\end{enumerate}

\paragraph{Generalizable Insights}
Beyond the specific application to kidney transplantation, this work provides broader insights into the role of machine learning in healthcare decision-making. A key challenge in deploying predictive models in clinical settings is bridging the gap between evaluation metrics and the decisions these models are intended to support. Many existing approaches evaluate models using population-level metrics that do not reflect how decisions are actually made in practice. 
In contrast, our paired recipient-based framework aligns evaluation with real-world allocation by examining the \emph{counterfactual benefit} of assigning a different recipient for a deceased donor's kidney.
This counterfactual perspective enables a more meaningful evaluation of model utility, moving beyond abstract prediction accuracy toward decision-relevant impact. More broadly, our approach highlights the importance of designing machine learning systems that are not only accurate, but also context-aware and aligned with the operational constraints of real-world deployment, particularly in high-stakes healthcare settings. 

\section{Background}

\subsection{Kidney Transplantation}
\label{sec:kidney_background}

\paragraph{Factors Affecting Time to Graft Failure}
The most successful kidney transplants last over the recipient's life span, but most transplants inevitably fail over time, and the time until graft failure or \emph{survival time} is affected by many different factors. 
Previous research shows that the age and race of both the donor and recipient are among the most significant factors in determining the graft survival time. 
The younger the donor and the older the recipient (up to about age 60), the longer the kidney graft tend to remain functional \cite{ashby2017kidney}. 
As for race, black recipients tend to achieve the shortest lasting kidney transplants, while Asian recipients are associated with the longest lasting kidney transplants \cite{gordon2010disparities}. 
Prior research has revealed many possible reasons why the recipient's race so strongly influences the graft survival time \cite{fan2010access}. 

The ischemia/reperfusion (IR) time of the kidney leading up to transplantation affects its survival time as well. With newest technology of continuous perfusion, kidneys can safely undergo cold ischemia for 24 hours and potentially even longer, but shorter IR time benefits the quality of kidney transplants \cite{ponticelli2015impact}. 
Another important factor is the disparity of the Human Leukocyte Antigens (HLAs) between a donor and a recipient, which affects the potency of the recipient's immune response towards the transplanted kidney.
A greater number of HLA mismatches (MMs) is associated with shorter graft survival times, particularly for HLA-DR locus MMs \cite{opelz1999hla, ashby2017kidney}. 
Incorporating feature representations for HLA compatibility, both at the level of MMs and serological HLA types, into survival prediction models has also been found to improve predictions of graft survival time \cite{nemati2023predicting}.

\paragraph{Donor and Recipient Matching}
Current organ allocation practice strives to balance three main metrics \cite{lee2019allocation}: fairness, by offering kidneys to patients who waited for the longest time; special consideration in the priority for highly sensitized patients; and matching the 20\% of best donors and 20\% of best recipients with the longest predicted survival times, called longevity matching \cite{asfour2024association}. Both the KDPI and EPTS are employed for improvement in donor-recipient matching. However, the metrics for quantification of donor quality of KDPI had limited predictive accuracy \cite{Molinari2022HighKDPI}. Death-censored graft survival was similar among different KDPI groups.

Many alternatives for the current allocation policies are possible. For example, one could consider the top two or more candidates and choose the one with the longest predicted time to graft failure according to a survival prediction model, thus maximizing utility, although potentially at the cost of lower fairness.
Any policy decisions must therefore be grounded in robust evidence, as they have profound implications, not only for the recipient but also for others awaiting transplants. 

\subsection{Survival Prediction}
Survival prediction models are designed to estimate the probability that an event of interest, such as graft failure or patient death, will occur over time. Unlike traditional classification or regression problems, survival prediction must account for censoring, which occurs when the event has not yet been observed for some subjects during the study period.

\paragraph{Survival Prediction Models}
A broad range of survival prediction models have been developed. 
We briefly describe the models we consider in this paper, selected from prior studies on survival prediction for kidney transplantation, which we further discuss in Section \ref{sec:related}. 
The Cox Proportional Hazards (CoxPH) model is one of the most widely used survival models \cite{cox1972regression}. It estimates the hazard (risk of event at a time point) as a function of patient features, assuming that these effects are constant over time (proportional hazards).
Coxnet is a regularized version of the CoxPH model that adds L1 (lasso) and L2 (ridge) penalties to prevent overfitting and perform automatic feature selection \cite{simon2011regularization}.
DeepSurv extends CoxPH using a neural network to model nonlinear effects while preserving the proportional hazards assumption \cite{katzman2018deepsurv}. 

Beyond the CoxPH model and its extensions, 
random survival forests (RSFs) have been proposed as an extension of random forests to censored data \cite{ishwaran2008random}. 
It builds an ensemble of decision trees, each trained on a different subset of the data, and aggregates their predictions. 
Another alternative is multi-task logistic regression (MTLR), which treats survival prediction as a sequence of binary classification tasks over time intervals \cite{yu2011learning}. It estimates the probability of survival up to each time point using a set of logistic regressions, which are trained together. 
The neural multi-task logistic regression (N-MTLR) model builds upon the MTLR formulation, but is powered by a deep neural network architecture \cite{fotso2018deepneuralnetworkssurvival}. 
Lastly, DeepHit is another deep learning-based survival model that directly learns the probability distribution over event times. It uses a loss function that combines likelihood with ranking accuracy \cite{lee2018deephit}. 

\paragraph{Evaluation Metrics}
Evaluating the performance of survival prediction models is critical to understanding their clinical utility, particularly in the high-stakes context of kidney transplantation.
One of the most widely used metrics is the concordance index (C-index), which measures a model's ability to correctly rank individuals by their risk or predicted survival time \cite{harrell1982evaluating}. Specifically, the C-index evaluates whether, for a randomly selected pair of patients, the one who experienced the event earlier also had a worse predicted outcome. A C-index of 1 indicates perfect discrimination, while a value of 0.5 suggests no better performance than random guessing. Its popularity is reflected in the literature: \citet{zhou2023survmetrics} report that over 80\% of survival prediction studies published in leading
statistical journals in 2021 use the C-index as their primary evaluation metric.

Recent critiques have highlighted additional shortcomings of the C-index beyond its mismatch with practical usage. As argued by \citet{lillelund2025stop}, the C-index measures only discriminative ability (i.e., ranking correctness), and does not assess calibration or the accuracy of predicted event times, making it a partial and potentially misleading evaluation metric for survival models.
We discuss these limitations further in the context of kidney transplantation in Section \ref{sec:related}.

\subsection{Related Work}
\label{sec:related}

\paragraph{ML-based Kidney Transplant Survival Prediction}
A recent systematic review by \citet{vandeKlundert2025comparative} comprehensively evaluated the landscape of kidney transplant survival prediction models, analyzing 37 studies and 134 comparative experiments published between 2010 and 2023. The review found that the majority of studies focused on predicting graft survival used models such as CoxPH, RSF, boosting, and support vector machines.
Several other approaches first divide recipients into subgroups and then train different models for different subgroups.
\citet{mark2019kidney} developed an ensemble model using RSF for one subgroup and CoxPH for another subgroup.
\citet{zhang2023multi} proposed an algorithm that 
uses MTLR within subgroups to identify risk factors and estimate survival probabilities.

Recent comparisons between ML-based survival prediction models, including deep neural network-based models, and traditional statistical models such as the CoxPH model, have yielded mixed results. 
\citet{paquette2022kidney} evaluated a variety of survival prediction models for graft survival, including CoxPH, DeepSurv, and DeepHit and found that neural network models outperformed traditional methods in discriminative ability. 
On the other hand, the findings of \citet{Truchot2023MLvsStats} 
suggest that ML models did not consistently outperform traditional approaches such as CoxPH, particularly in terms of calibration and external validity. 
These results reinforce the continued relevance of CoxPH-based methods such as Coxnet, which we include in our own model comparisons.

\paragraph{Evaluation Metrics for Kidney Transplant Survival Prediction}
The recent review by \citet{vandeKlundert2025comparative} categorizes prediction performance metrics into three key dimensions: calibration, discrimination, and classification. While each offers valuable insight into different aspects of model behavior, discrimination, which assesses how well a model can rank individuals by risk, was by far the most commonly reported dimension. Specifically, the C-index was used in 39 out of the 44 studies with clearly reported metrics. This metric measures the concordance between predicted and actual event orderings.

However, despite its widespread adoption, the C-index has important limitations, particularly in the context of organ allocation. The standard C-index formulation evaluates pairwise concordance across all possible recipient pairs in the dataset, regardless of whether those recipients received kidneys from different donors, in different years, or under very different clinical circumstances. In practical allocation scenarios, however, decisions are donor-specific: clinicians must choose between a small pool of candidates to identify the best recipient for a particular available donor kidney. As such, evaluating whether a model correctly orders unrelated recipients from across the dataset does not reflect how it would be used in practice.

\citet{wolfe2009predictability} make another important point: the maximal value of the C-index is constrained by the information available in the predictors and by the set of pairs being compared. They illustrate this with a simple example in which two diagnostic groups are perfectly separable from each other but indistinguishable within-group; a model built on diagnosis alone will be correct for every cross-group comparison but only correct by chance for within-group pairs, yielding a C-index of 0.75 even though the model is perfectly useful for clinically relevant distinctions. In other words, the C-index averages performance over many pairwise comparisons that may be uninformative, and can therefore understate a model's utility for the comparisons clinicians actually care about.

Given its prevalence and relevance to clinical decision-making, our study also emphasizes discrimination performance as the main axis of model evaluation. We report the C-index for graft survival time predictions to enable comparisons with prior work. However, we go further by developing practical, decision-informed metrics that directly reflect the prediction model's utility in the context of organ allocation, aligning predictive accuracy with clinical outcomes. In doing so, we take a step to address the need for more actionable and externally valid evaluation frameworks.

\section{Materials and Methods}

\subsection{Data Description}
This study used data from the Scientific Registry of Transplant Recipients (SRTR). The SRTR data system includes data on all donor, wait-listed candidates, and transplant recipients in the US, submitted by the members of the Organ Procurement and Transplantation Network (OPTN). The Health Resources and Services Administration (HRSA), U.S.~Department of Health and Human Services provides oversight to the activities of the OPTN and SRTR contractors.

The prediction models in this study were developed using records from a total of 111,518 deceased donor kidney transplants performed between the years 2000 and 2016. 
The prediction target in this study is death-censored graft loss, which is defined as graft failure not attributable to recipient death. This formulation ensures that graft loss is treated independently of patient mortality. A graft is considered lost if there is a recorded instance of graft failure, return to maintenance dialysis, re-transplantation, or listing for re-transplantation. If none of these events occur and the recipient dies with a functioning graft, the outcome is treated as censored. For all censored instances, the censoring time is defined as the last known follow-up date.

The features we use to predict transplant outcomes include fundamental clinical and demographic features such as donor and recipient age, body mass index (BMI), gender, and race. 
We also include the total number of HLA MMs (0-6) between the donor and recipient for the HLA-A, -B, and -DR loci. 
All features used for prediction are \emph{pre-transplant} features, meaning that they are available prior to the transplant time and can be used for kidney allocation. 
We deliberately exclude post-transplant features such as the cold ischemia time and the recipient serum creatinine at
discharge time, which have been found to improve prediction accuracy \cite{nemati2023predicting}, but are not available prior to the transplant time.

\subsection{Proposed Paired Recipient-based Evaluation}
To address the disconnect between current C-index based evaluations and kidney allocation, we propose a new evaluation method tailored to the retrospective nature of our dataset. In most cases, each deceased donor contributed two kidneys that were \emph{transplanted into two distinct recipients}. Since the donor is the same for both recipients, this natural pairing allows us to evaluate models in a way that more closely mirrors real-world decision-making. Our proposed metric assesses whether the model correctly identifies, between the two recipients of kidneys from the same donor, which one ultimately experienced the longer graft survival. This pairwise donor-based evaluation aligns more directly with the actual clinical objective: choosing the more compatible recipient for a given donor kidney to enable a longer lasting graft. Our proposed evaluation approach is illustrated in Figure~\ref{fig:eval_method}.

\subsubsection{Comparable Recipient Pairs}

To support our proposed pairwise evaluation method, we distinguish \emph{comparable recipient pairs} from the rest of the dataset. In this context, a recipient pair refers to the 2 recipients who received kidneys from the same deceased donor. Our evaluation goal is to determine whether the model correctly identifies which recipient in each pair had the longer-lasting graft. However, due to the presence of censoring, not all pairs provide a definitive ground truth for graft survival comparison.

Censoring occurs when the event of interest (in this case, graft failure) has not been observed during the follow-up period. This can happen for several reasons: the graft may still be functioning at the time of data collection (the most common scenario); the patient may have died with a functioning graft, thus precluding the observation of the actual graft failure time; or the patient was lost in later follow ups. Since censoring obscures the true event time, not all recipient pairs are usable for evaluating which recipient had the longer-lasting graft.

\begin{figure}[t]
  \centering
  \floatconts{fig:comparable_cindex_diff}%
  {\caption[Comparable and incomparable recipient pairs. 
  Difference between pairwise comparisons in our paired recipient-based evaluation and pairwise comparisons used in computing the C-index.]{\subfigref{fig:comparable} Comparable and incomparable recipient pairs. 
  \subfigref{fig:cindex_diff} Difference between pairwise comparisons in our paired recipient-based evaluation and pairwise comparisons used in computing the C-index.}}%
  {%
  \subfigure{%
    \includegraphics[width=0.4\textwidth]{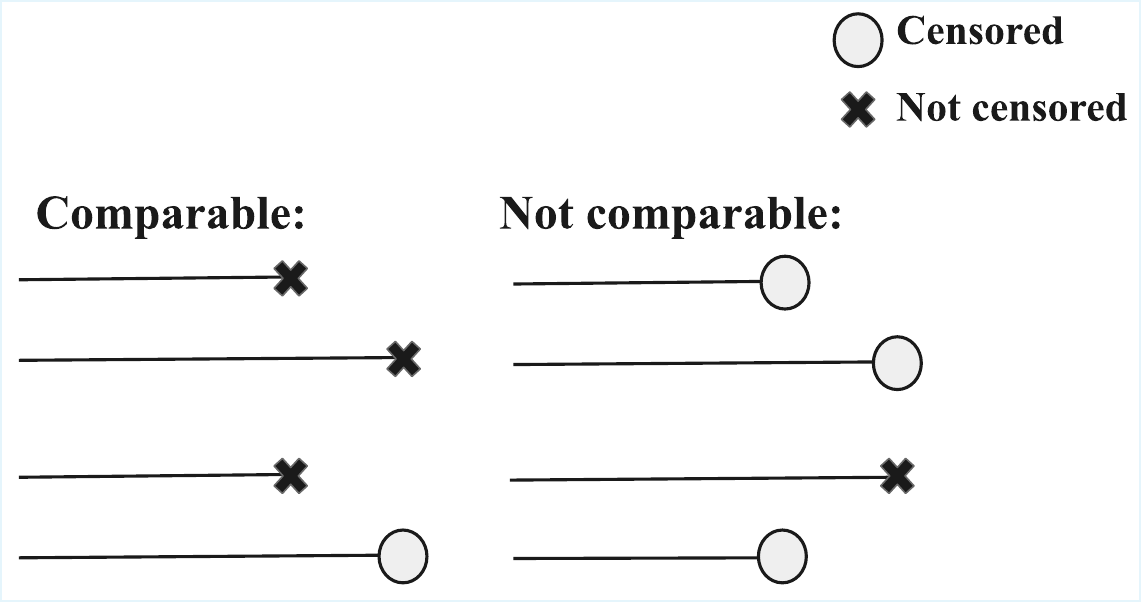}
    \label{fig:comparable}
  }
  \hfill
  \subfigure{%
    \includegraphics[width=0.57\textwidth]{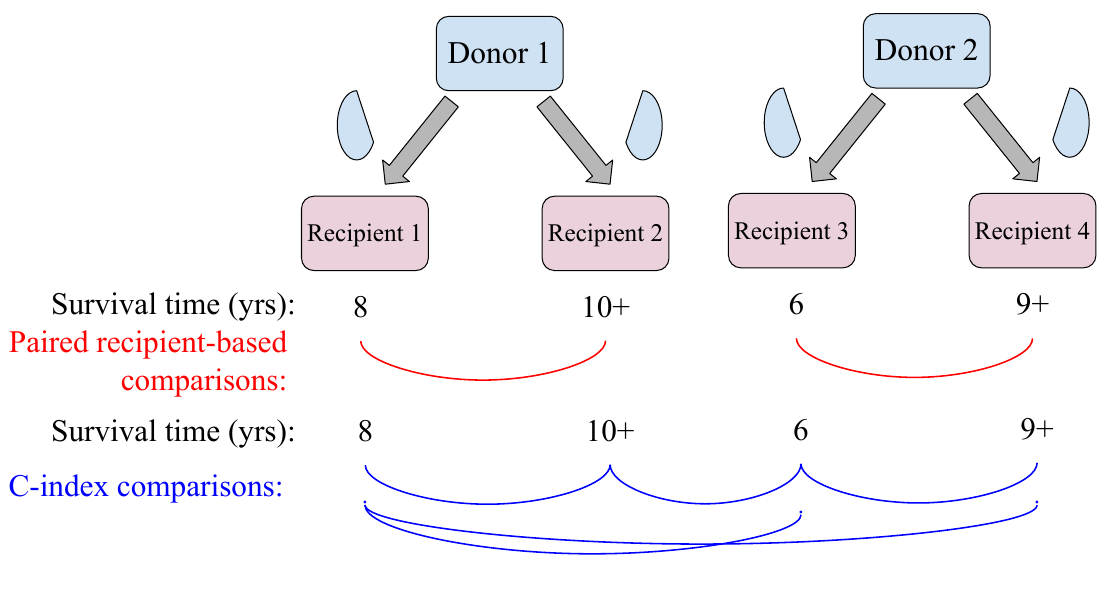}
    \label{fig:cindex_diff}
  }
  }
\end{figure}

We define a recipient pair as \emph{comparable} if we observe which of the two grafts survived longer. This condition is met under two scenarios, illustrated in Figure~\ref{fig:comparable}:
\begin{enumerate}
\item Both recipients were uncensored, meaning graft failure was observed for both, allowing for a direct comparison of survival durations.
\item One recipient was censored, but their censored survival time was longer than the observed graft failure time of the other recipient. In this case, we can reasonably infer that the censored graft outlasted the failed one, even if we do not know its exact time of failure.
\end{enumerate}

Out of the 55,759 donors in our dataset (i.e., 111,518 transplants), 11,952 pairs were comparable recipient pairs while the remaining 43,807 pairs were incomparable. 

\subsubsection{Differences from Current Evaluation Approaches}
Figure~\ref{fig:cindex_diff} illustrates the difference between our proposed paired recipient-based evaluation method and the commonly used C-index, which also uses pairwise comparisons. In contrast to the C-index, our proposed paired recipient-based evaluation metric is donor-specific. It focuses exclusively on recipient pairs who received kidneys from the same deceased donor, which is a much closer match to how transplant decisions are made in practice. When a kidney becomes available, the goal is not to predict which recipient in the entire dataset (many who have already received a transplant and are no longer on the waiting list) would have the longest survival, but to choose between a small set of candidates on the waiting list who would be available for a kidney from a specific donor. Thus, our paired recipient-based accuracy metric avoids inappropriate comparisons between recipients with unrelated donor characteristics that may confound the model's evaluation.

A key advantage of our proposed paired recipient-based evaluation is that it enables estimation of the \emph{counterfactual benefit} of changing kidney allocation policy. 
Typical evaluations of changes in organ allocation policies require simulated allocation models (SAMs), which have a long history in organ transplantation \cite{cremers2026global}. 
A key weakness of SAMs is that they require ``submodels'' to estimate patient and graft survival when a donor's organ is transplanted to a recipient. 
Hence, they use simulated outcomes (because the transplant never actually occurred) to evaluate the counterfactual benefit from changes in allocation policies. 
On the other hand, our proposed evaluation uses \emph{actual outcomes}, as we compare between two transplants that actually occurred. 
This enables a more reliable evaluation of changes in allocation policies, both in terms of utility and fairness, which we describe in the following.

\subsubsection{Post-transplant Years Gained}
While predictive accuracy measures a model's ability to correctly identify which recipient in a same-donor pair had longer graft survival, it does not by itself indicate the clinical benefit of using the model for decision-making. To improve interpretability, we also report \emph{post-transplant years gained}. This metric estimates the additional years of graft survival that would be expected if the recipient was selected using the survival prediction model rather than by randomly selecting one of the two recipients.

To compute this metric, we first assume a baseline model with 50 percent accuracy, which represents random selection between the two recipients from the same donor. We then calculate the difference in average survival time between the grafts predicted by the model to last longer and those that would have been selected randomly. This quantity provides a tangible estimate of how many additional years of graft function a model could contribute if used in real-world donor-recipient matching. We note that the baseline used as comparison here represents random choice within observed same-donor recipient pairs, which should not be interpreted as a simulation of the current organ allocation policy as it does not take into account other factors such as equity, which we discuss in Section \ref{sec:fairness}.

Since many recipients in our test set are censored, their actual graft survival time is unknown. To estimate survival time for these cases, we apply a 
trained CoxPH model
to generate individualized conditional survival curves \cite{haider2020effective} for the censored test recipients. For each censored recipient, we estimate their expected graft survival time by identifying the time point at which their predicted survival probability falls to 0.5 (i.e., their median predicted survival time).
To account for uncertainty in these estimates, we also compute an interquartile range (IQR). This is done by identifying the survival durations corresponding to survival probabilities of 0.75 and 0.25, which give the 25th and 75th percentile survival times, respectively.

Our approach differs from the Life Years from Transplant (LYFT) framework proposed by \citet{wolfe2008lyft}, despite the similar name. 
LYFT estimates the additional years of life a candidate can expect from transplantation compared to remaining on dialysis. LYFT includes both time with a functioning graft and time after graft failure, focusing on recipient life longevity. While it uses similar survival modeling techniques, our goal is to estimate the extra years of graft function gained by using predictive models to guide donor-recipient matching. This shifts the focus from individual life expectancy to system-level efficiency, aiming to reduce re-transplantation rates and make better use of available organs. Although longer graft survival can contribute to longer patient lives, our metric is centered on maximizing graft utility, not life years.

\subsubsection{Fairness across Racial Groups}
\label{sec:fairness}
In addition to predictive accuracy and post-transplant graft years gained, we evaluate the fairness across racial groups, if a survival prediction model is used to select the recipient. 
As described in Section \ref{sec:kidney_background}, prior research has found that graft survival time is highly-dependent on the recipient's race.
We focus on the four most represented racial categories in our dataset: Asian, Black, Hispanic, and White recipients.

We adopt \emph{demographic parity} as our fairness metric. Let $\hat{Y} \in \{0,1\}$ denote the allocation decision, where $\hat{Y}=1$ indicates that a candidate is selected to receive the donor kidney, and let $A$ denote the protected attribute (race). Demographic parity requires that the probability of selection be independent of race:
\begin{equation*}
P(\hat{Y} = 1 \mid A = a) = P(\hat{Y} = 1 \mid A = b)
\quad \forall a,b.
\end{equation*}

To measure violation of demographic parity, we use the \emph{difference of demographic parity} \cite{lei2024inductivebiasesdemographicparitybased}. In our paired allocation setting, each donor kidney is assigned to one of two candidates. Under random selection, each candidate would be chosen with probability 0.5. Therefore, for each racial group $a$, we compute the observed selection rate:
\begin{equation*}
\text{DP}(a) = P(\hat{Y} = 1 \mid A = a),
\end{equation*}
and report its deviation from 0.5, which is the probability that a candidate is selected by random chance:
\begin{equation*}
\Delta(a) = \left| \text{DP}(a) - 0.5 \right|.
\end{equation*}

A value of $\Delta(a)=0$ indicates perfect demographic parity under random baseline expectations, while larger values reflect greater imbalance in how frequently candidates from group $a$ are selected. 
We denote the total imbalance by $\sum \Delta$, the sum over all racial groups.

From a fairness perspective, demographic parity promotes equal access to transplantation across racial groups. However, maximizing graft survival may conflict with this objective, since prior research has shown that post-transplant outcomes are statistically associated with demographic variables, including race. A purely utility-driven model may therefore disproportionately favor groups with historically better observed outcomes.

We choose demographic parity as our fairness metric for three reasons. First, it is straightforward to compute and interpret, making it accessible to clinicians, policymakers, and the general public. Second, it aligns naturally with our paired allocation framework, where equal selection probability (0.5) provides a clear baseline reference. Third, while more complex fairness definitions exist (e.g., equalized odds or calibration-based criteria), demographic parity directly captures disparities in allocation rates, which are central to the ethical and societal acceptability of organ allocation policies.

To provide context for the fairness analysis, Table~\ref{tab:race_dist} summarizes the distribution of recipients by race in the full dataset and in the subset of comparable recipient pairs used for evaluation. 
We find that black recipients are overrepresented in the subset of comparable recipient pairs, while other races are slightly underrepresented. 
The overrepresentation of black recipients is likely due to their shorter graft survival times \cite{gordon2010disparities}, leading to less incomparable pairs where both recipients are censored, or one recipient is censored before the failure time of the other recipient's graft. 

\begin{table}[t]
\centering
\caption{Distribution of recipients by race in the full dataset and in the subset of comparable recipient pairs. Percentages are reported within each group. 
The total percentages do not add up to 100\% because some extremely small racial groups such as native Hawaiian or Pacific islander are excluded.}
\label{tab:race_dist}
\begin{tabular}{lcccc}
\toprule
\textbf{Race} & \textbf{Full Dataset} & \textbf{(\%)} & \textbf{Comparable Subset} & \textbf{(\%)} \\
\midrule
Asian    & 6,799  & 6.2\%  & 1,221  & 5.2\% \\
Black    & 35,666 & 32.7\% & 9,178  & 39.2\% \\
Hispanic & 17,064 & 15.6\% & 3,273  & 14.0\% \\
White    & 49,819 & 45.6\% & 9,756  & 41.6\% \\
\midrule
Total    & 109,348& 98.1\%  & 23,428 & 98.0\% \\
\bottomrule
\end{tabular}
\end{table}

\subsection{Survival Prediction Algorithms}

\subsubsection{Machine Learning-based Predictors}
\label{sec:ml_predictors}
We evaluate five machine learning-based survival prediction models commonly used in survival analysis: Coxnet, DeepSurv, DeepHit, Neural Multi-Task Logistic Regression (N-MTLR), and Random Survival Forest (RSF). 
We adopt a nested $5\times 2$ cross-validation (CV) strategy over the full dataset. 
We first randomly partition the entire dataset into five outer folds, using donor ID as the grouping variable to ensure that both recipients from each pair are assigned to the same fold and therefore never split across training and test sets to avoid leakage of donor information. In each outer CV iteration, one fold is held out as the test set, while the remaining four folds are used for model training.
Within the training folds, we perform an inner two-fold CV to tune model-specific hyperparameters. 
Once optimal hyperparameters are selected, the models are retrained on the full training folds from the outer CV and evaluated on the corresponding outer CV test fold. 
The hyperparameter values we consider for each model are discussed in Appendix \ref{sec:hp_tuning}. 

For each test fold, we compute multiple evaluation metrics reflecting different aspects of model performance. Our proposed paired recipient-based accuracy and the associated post-transplant graft years gained are calculated only on comparable recipient pairs within the test fold, using the same comparability criteria defined earlier in this section. In contrast, the concordance index (C-index) is computed over the entire test fold, following the standard formulation used in prior work.
This process is repeated across all five outer test folds. Final reported results represent the mean and standard error across CV folds for each metric.

\subsubsection{Single Factor-based Models}
In addition to the multi-feature machine learning-based survival prediction models, we also evaluate a set of simple rule-based allocations based on a single factor. These policies simulate a scenario in which a clinician chooses between two transplant candidates based solely on one predefined criterion, ignoring all other variables.
These single factor-based predictors provide an interpretable baseline and allow us to assess the marginal impact of individual factors on graft survival. We focus separately on age, race, and HLA mismatches, resulting in the following decision policies:
\begin{itemize}
    \item Recipient Age: choose the older recipient.
    \item Recipient Race: choose the recipient by race in the following order: Asian $>$ Hispanic $>$ White $>$ Black.
    \item HLA MM: choose the recipient with lower HLA MM with the donor (lower DR MM in case of tie).
\end{itemize}

The selection order for race is determined based on their impact on graft survival, as indicated by the CoxPH coefficients for each binary race variable. 
These policies are applied to all comparable recipient pairs, with outcomes measured using the same metrics as the ML-based models: paired recipient-based prediction accuracy, C-index, and post-transplant years gained.

\section{Results}

\subsection{Utility}
\begin{table}[t]
\centering
\caption{Accuracy of ML-based survival prediction models. Results are reported as mean $\pm$ standard error (SE). Post-transplant years gained are reported at the 25th, 50th, and 75th percentiles. ML-based models are on top, while single factor-based models are on the bottom.}
\label{tab:results}
\begin{tabular}{lccccc}
\toprule
& \textbf{Paired} & & \multicolumn{3}{c}{\textbf{Post-transplant Years Gained}}\\
\textbf{Model} & \textbf{Accuracy} & \textbf{C-index} & \textbf{25\%} & \textbf{50\%} & \textbf{75\%} \\
\midrule

Coxnet   & $0.598 \pm 0.005$ & $0.632 \pm 0.002$ & $1.48 \pm 0.06$ & $2.06 \pm 0.07$ & $2.65 \pm 0.10$ \\
DeepHit  & $0.599 \pm 0.008$ & $0.625 \pm 0.002$ & $1.46 \pm 0.07$ & $2.03 \pm 0.08$ & $2.60 \pm 0.12$ \\
DeepSurv & $0.599 \pm 0.007$ & $0.632 \pm 0.002$ & $1.51 \pm 0.06$ & $2.10 \pm 0.07$ & $2.71 \pm 0.11$ \\
N-MTLR   & $0.594 \pm 0.004$ & $0.628 \pm 0.001$ & $1.41 \pm 0.04$ & $1.94 \pm 0.04$ & $2.55 \pm 0.06$ \\
RSF      & $0.601 \pm 0.005$ & $0.631 \pm 0.002$ & $1.48 \pm 0.06$ & $2.06 \pm 0.07$ & $2.65 \pm 0.11$ \\
\midrule
Recipient Age & $0.570 \pm 0.003$ & $0.553 \pm 0.001$ & $1.12 \pm 0.02$ & $1.62 \pm 0.02$ & $1.99 \pm 0.04$ \\
Recipient Race & $0.553 \pm 0.002$ & $0.567 \pm 0.001$ & $0.82 \pm 0.02$ & $1.09 \pm 0.03$ & $1.44 \pm 0.03$ \\
HLA MM & $0.526 \pm 0.006$ & $0.546 \pm 0.002$ & $0.34 \pm 0.06$ & $0.48 \pm 0.08$ & $0.61 \pm 0.11$ \\
\bottomrule
\end{tabular}
\end{table}

Table~\ref{tab:results} summarizes the performance of the survival prediction models using three evaluation metrics for utility: paired recipient-based accuracy, C-index, and the 25th, 50th, and 75th percentiles for post-transplant years gained per transplant. 
We find that all five ML-based models achieve roughly the same paired recipient-based accuracy of about 60\%. 
Recall that the paired recipient-based accuracy reflects how often a model correctly predicts which of the two recipients for the same donor had the longer-lasting graft. 

The ML-based models are all more accurate than the single factor-based models, which range from 52-57\% accuracy. 
The results for single factor-based models are consistent with prior clinical knowledge. For instance, older recipients are known to have more tolerant-prone immune responses to transplanted organs \cite{ashby2017kidney}, and Asian recipients generally show favorable post-transplant outcomes compared to Black and Caucasian recipients. Conversely, Black recipients perform worse than other races even if offered preferential donor selection \cite{gordon2010disparities, Bekbolsynov2022HLAImmunogenicity}.

The C-index shows slightly more variation across the ML-based models, ranging from $0.625$ for DeepHit to $0.632$ for Coxnet and DeepSurv. 
Again, the C-indices are lower for the single factor-based models, ranging from $0.546$ for HLA MM to $0.567$ for Recipient Race. 
Unlike our pairwise evaluation, which compares outcomes only between recipients of the same donor, the general C-index compares survival predictions across all possible recipient pairs, including those from different donors, different decades, and different clinical contexts. 
As such, the C-index may overestimate the models' ability to identify transplants with higher risk by taking advantage of the variability in donor-related features that are irrelevant to kidney allocation decisions when a specific donor kidney is available. 

We find that the rank order of the models in terms of paired recipient-based accuracy may disagree with that of the C-index.
For example, N-MTLR has a lower paired recipient-based accuracy than DeepHit, but a higher C-index. 
This may indicate that N-MTLR can more accurately rank transplants with different donors compared to DeepHit, while performing worse at predicting which of the two recipients from the same donor will have the better outcome. 
The same observation applies to Recipient Race vs.~Recipient Age for the single factor-based models.

The third metric, mean post-transplant years gained per transplant, offers an interpretable way to quantify the counterfactual benefit of using survival models for recipient selection. This value represents the additional years of graft function expected when using model-guided allocation instead of a random choice. 
Similar to the paired recipient-based accuracy, we find that all of the ML-based models achieve roughly the same number of post-transplant years gained, with a median of around 2 years gained per transplant. 
In comparison, the single factor-based models can achieve a median of about 0.5-1.6 years gained per transplant, suggesting that the improved prediction accuracy does translate into a non-negligible improvement in graft survival times.

\subsection{Fairness}

\begin{table}[t]
\setlength{\tabcolsep}{5.5pt}
\centering
\caption{Demographic parity (DP) by race and total deviation from parity ($\sum \Delta$) for each model.}
\label{tab:dp}
\begin{tabular}{lccccc}
\toprule
\textbf{Model} & \textbf{Asian} & \textbf{Black} & \textbf{Hispanic} & \textbf{White} & $\boldsymbol{\sum \Delta}$ \\
\midrule

Coxnet   & $0.680 \pm 0.008$ & $0.300 \pm 0.003$ & $0.557 \pm 0.006$ & $0.647 \pm 0.004$ & $0.584 \pm 0.012$ \\
DeepHit  & $0.657 \pm 0.015$ & $0.323 \pm 0.003$ & $0.543 \pm 0.005$ & $0.634 \pm 0.003$ & $0.510 \pm 0.018$ \\
DeepSurv & $0.746 \pm 0.009$ & $0.295 \pm 0.005$ & $0.573 \pm 0.015$ & $0.642 \pm 0.010$ & $0.665 \pm 0.011$ \\
N-MTLR   & $0.714 \pm 0.013$ & $0.303 \pm 0.004$ & $0.561 \pm 0.011$ & $0.643 \pm 0.003$ & $0.615 \pm 0.015$ \\
RSF      & $0.633 \pm 0.024$ & $0.306 \pm 0.002$ & $0.574 \pm 0.004$ & $0.644 \pm 0.003$ & $0.545 \pm 0.018$ \\

\bottomrule
\end{tabular}
\end{table}

Table~\ref{tab:dp} reports demographic parity (DP) across racial groups along with the total deviation from parity, $\sum \Delta$, for each model. Consistent with earlier observations, all models exhibit noticeable disparities in selection rates across races. Asian and White recipients are selected more frequently than the baseline of 0.5, while Black recipients are consistently selected at substantially lower rates across all models. Hispanic recipients tend to be closest to parity, with selection rates near 0.5.

The aggregated deviation metric $\sum \Delta$ provides a concise measure of overall fairness. Note that this value can range from 0 (most fair with all four races selected at a rate of 0.5) to 2 (least fair with all four races selected at a rate of 0 or 1). Among the models, DeepHit achieves the lowest total deviation (0.51), indicating the most balanced allocation across racial groups, while DeepSurv shows the largest deviation (0.66), reflecting greater unfairness. 
DeepSurv favors Asian recipients, choosing them at a much higher rate than the other models.

These results highlight that even when predictive performance is similar across models, their fairness properties can differ. In particular, models that achieve strong predictive accuracy do not necessarily produce equitable allocation outcomes. This underscores the importance of evaluating fairness alongside traditional performance metrics, as optimizing solely for graft survival may inadvertently favor certain demographic groups and exacerbate disparities in access to transplantation.

\section{Discussion}

The results of this study highlight the potential of survival prediction models, trained solely on pre-transplant data, to improve donor-recipient matching decisions in kidney transplantation. Using a novel paired recipient-based evaluation framework, we demonstrate that even modestly accurate models ($\sim 60\%$ paired recipient-based accuracy) can lead to clinically meaningful gains in graft survival compared to random selection, typically on the order of 1.5 to 2.5 years per transplant. At a societal level, about 21,000 deceased donor kidney transplants were performed in the United States in 2025 \cite{hrsa_unos_national_data}. A gain of 1.5-2.5 years per graft would translate to about 31,500-52,500 additional graft survival years annually. 
These gains could translate directly into improved organ utilization and reduced need for re-transplantation, which are both critical goals in transplant medicine. 
Given that the mean graft survival time for deceased donor kidney transplants is about 12 years \cite{poggio2021longterm}, this could result in an additional 2,600-4,300 transplants per year without increasing the number of deceased donors!

To further contextualize these results, we pose a hypothetical question. 
If we could increase prediction accuracy from 60\% to 65\%, what would be the improvement in post-transplant years gained? 
The answer depends on which 65\% of predictions are correct, as achieving a correct prediction on an extremely long-lasting graft (e.g., 20 years) yields a greater improvement than on a graft that lasts only a few years. 

\begin{figure}[t]
  \centering
  \floatconts{fig:simulated_gains}%
  {\caption[Simulation of graft years gained per transplant for different predictor accuracy levels, correct cases chosen randomly and strategically.]{Simulation of graft years gained per transplant for different predictor accuracy levels, correct cases chosen \subfigref{fig:simulation} randomly and \subfigref{fig:strategic_simulation} strategically.}}%
  {%
  \subfigure[Random predictor]{%
    \includegraphics[width=0.48\textwidth]{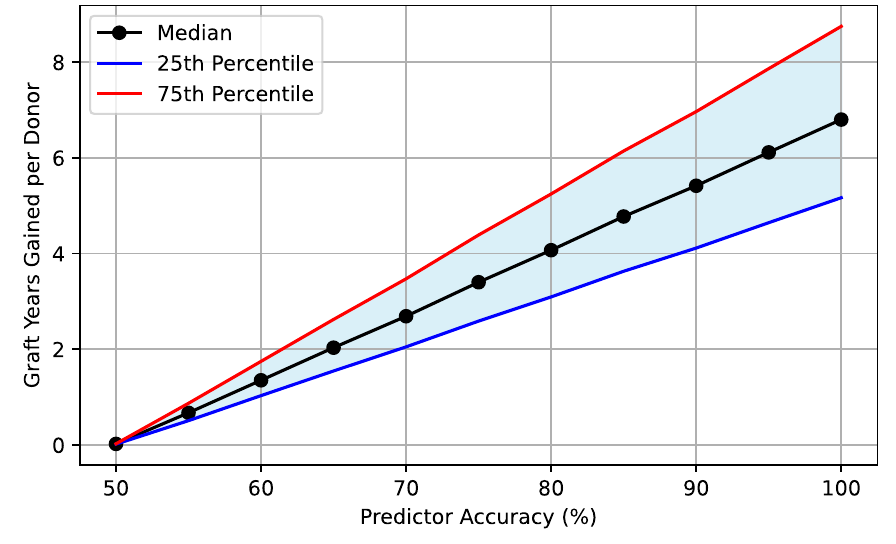}
    \label{fig:simulation}
  }
  \hfill
  \subfigure[Strategic (oracle) predictor]{%
    \includegraphics[width=0.48\textwidth]{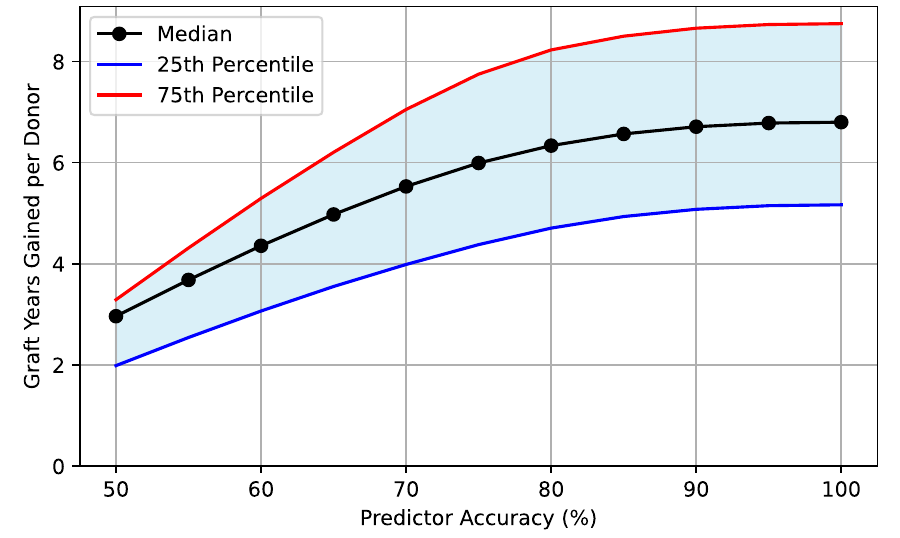}
    \label{fig:strategic_simulation}
  }
  }
\end{figure}

To arrive at a pessimistic answer, we assume that the predictor randomly selects the 65\% of correct predictions and conduct a simulation that quantifies expected graft years gained per transplant as a function of predictor accuracy, ranging from 50\% (random choice) to 100\% (perfect accuracy). Each point in Figure~\ref{fig:simulation} represents the average years gained across 100 random predictors at a given accuracy level, with shaded confidence bands denoting the interquartile range (25th to 75th percentile).
The simulation shows a strong linear relationship, with expected graft years gained per transplant increasing steadily from 0 years at 50\% accuracy to approximately 5.2-8.8 years at 100\% accuracy. This provides a reference for interpreting the performance of our models and future models. 
From Table \ref{tab:results}, we find that the ML-based predictors achieve about 60\% accuracy and a median post-transplant gain of 2 years, which is higher than the 1.5 years expected from a random predictor that achieves 60\% accuracy.

In addition to the simulation of a random predictor, we consider a strategic predictor that preferentially allocates its correct predictions to the donor-recipient pairs with the largest potential graft year differences to arrive at an optimistic prediction. 
This can also be considered as an oracle prediction, as it requires knowledge of the transplant outcomes and is not achievable in practice. 
The results are shown in Figure~\ref{fig:strategic_simulation}.
Unlike the linear trend observed in the random setting, the strategic predictor shows a nonlinear curve with diminishing returns as accuracy increases. Even at 50\% accuracy, this approach yields an additional 2-3.5 graft years per transplant compared to random allocation, since the correct predictions are concentrated on the longest lasting transplants compared to the other recipient from the same donor. 
This optimistic prediction indicates that the accuracy itself is not the only relevant factor, but rather, correctly choosing the recipients who will achieve the best outcomes and avoiding those who will have the worst outcomes. 

\section{Conclusion}
Our main focus in this study was to evaluate survival prediction accuracy for deceased donor kidney transplants in a clinically relevant and actionable manner. 
To achieve this, we proposed several paired recipient-based evaluation metrics that can directly translate to kidney allocation policy. 
Using these metrics, we found that five published survival models achieved similar paired recipient-based accuracy, all around 60\%. 
If the models were to be used to select between the two recipients of kidneys from the same deceased donor, this would yield an additional 1.5-2.5 years per transplant compared to random selection within the pair of recipients. 

Our results also reinforce limitations of the commonly used C-index in this setting. In both empirical and simulation analyses, the C-index fails to reflect donor-specific decision-making and can be misleading when comparing unrelated recipients. The paired recipient-based accuracy and associated post-transplant years gained offer a more relevant and interpretable evaluation framework for models intended to inform allocation decisions.

At the same time, our estimate of years gained has important limitations. First, it is defined relative to the random selection within previously observed recipient pairs, not the current kidney allocation policy. This baseline isolates the utility of pairwise survival discrimination, but it does not represent the real allocation process, which balances utility and equity. As a result, the estimated 1.5-2.5 year gain should not be interpreted as the expected improvement over current practice and would likely be smaller if incorporated into the current allocation system.

Second, our estimate of post-transplant years gained relies on pseudo-values derived from a CoxPH model. This inherits the assumptions of CoxPH, including proportional hazards and linear covariate effects on the log-hazard scale. In transplantation, where nonlinearities and complex donor-recipient interactions are plausible, these assumptions may be restrictive. 
If other models for generating individual survival curves and deriving pseudo-values \cite{haider2020effective} were used, our reported 1.5-2.5 years gained may differ.

Third, we focused our analysis on ML-based survival models that have been previously applied to kidney transplant data. 
Recent developments in ML research have yielded new survival prediction models, including attention-based models \cite{meng2022novel}, that have yielded superior prediction accuracy in other biomedical settings and could possibly improve beyond the roughly 60\% paired recipient-based accuracy that we observed in this paper.

Our findings also raise potential ethical concerns. Simple decision rules that improve average graft survival may still disadvantage certain groups, and our fairness analyses emphasize the risks of using sensitive attributes such as age or race in allocation decisions.

We see many interesting avenues for future work. 
First, our proposed paired recipient-based accuracy metric focused on discrimination ability, not calibration. Analyzing and improving calibration of ML-based survival prediction models could also be clinically relevant, particularly if used to predict post-transplant years gained. 
Secondly, evaluating counterfactual benefit against more complex baselines beyond randomly choosing one of the two recipients could provide more realistic estimates of post-transplant years gained if a new prediction model is incorporated into kidney allocation policy. 
Finally, we see tremendous value in incorporating fairness considerations to enable donor-recipient matches with longer lasting grafts while still maintaining a guarantee of equity.

\acks{The authors thank Nimra Gurung for her assistance in the data preparation.

This work made use of the High Performance Computing Resource in the Core Facility for Advanced Research Computing at Case Western Reserve University (CWRU) and was supported by a summer research scholarship provided by the CWRU Undergraduate Research Office.

Research reported in this publication was supported by the National Library of Medicine of the National Institutes of Health under Award Number R01LM013311 as part of the NSF/NLM Generalizable Data Science Methods for Biomedical Research Program. The content is solely the responsibility of the authors and does not necessarily represent the official views of the National Institutes of Health.

The data reported here have been supplied by the Hennepin Healthcare Research Institute (HHRI) as the contractor for the Scientific Registry of Transplant Recipients (SRTR). The interpretation and reporting of these data are the responsibility of the authors and in no way should be seen as an official policy of or interpretation by the SRTR or the U.S.~Government.%
}

\newpage
\bibliography{references}

\begin{thebibliography}{40}
\providecommand{\natexlab}[1]{#1}
\providecommand{\url}[1]{\texttt{#1}}
\expandafter\ifx\csname urlstyle\endcsname\relax
  \providecommand{\doi}[1]{doi: #1}\else
  \providecommand{\doi}{doi: \begingroup \urlstyle{rm}\Url}\fi

\bibitem[Adler et~al.(2021)Adler, Husain, King, and Mohan]{adler2021greater}
J.~T. Adler, S.~A. Husain, K.~L. King, and S.~Mohan.
\newblock Greater complexity and monitoring of the new kidney allocation system: Implications and unintended consequences of concentric circle kidney allocation on network complexity.
\newblock \emph{American Journal of Transplantation}, 21:\penalty0 2007--2013, 2021.
\newblock \doi{10.1111/ajt.16441}.

\bibitem[Al~Awadhi et~al.(2025)Al~Awadhi, Hsu, Potter, Kakadiaris, Axelrod, Parsons, Meinders, Cassell, Pulicken, Javed, Shireman, Casarin, Gelfond, and Waterman]{AlAwadhi2025MLDropout}
S.~Al~Awadhi, E.~Hsu, T.~B.~H. Potter, I.~A. Kakadiaris, D.~A. Axelrod, F.~Parsons, A.~M. Meinders, V.~Cassell, C.~Pulicken, Z.~Javed, P.~K. Shireman, S.~Casarin, A.~L.~J. Gelfond, and A.~D. Waterman.
\newblock Developing and validating machine learning-driven risk indices to predict patient dropout during referral, evaluation, and waitlisting for kidney transplant.
\newblock \emph{Clinical Transplantation}, 39\penalty0 (9):\penalty0 e70325, 2025.
\newblock \doi{10.1111/ctr.70325}.
\newblock URL \url{https://doi.org/10.1111/ctr.70325}.

\bibitem[Ali et~al.(2025)Ali, Shroff, Fülöp, Molnar, Sharif, Burke, Shroff, Briggs, and Krishnan]{Ali2025AITransplant}
H.~Ali, A.~Shroff, T.~Fülöp, M.~Z. Molnar, A.~Sharif, B.~Burke, S.~Shroff, D.~Briggs, and N.~Krishnan.
\newblock Artificial intelligence assisted risk prediction in organ transplantation: a uk live-donor kidney transplant outcome prediction tool.
\newblock \emph{Renal Failure}, 47\penalty0 (1):\penalty0 2431147, 2025.
\newblock \doi{10.1080/0886022X.2024.2431147}.
\newblock URL \url{https://doi.org/10.1080/0886022X.2024.2431147}.

\bibitem[Asfour et~al.(2024)Asfour, Zhang, Lu, Reese, Saunders, Peek, White, Persad, and Parker]{asfour2024association}
Nour~W. Asfour, Kevin~C. Zhang, Jessica Lu, Peter~P. Reese, Milda Saunders, Monica Peek, Molly White, Govind Persad, and William~F. Parker.
\newblock Association of race and ethnicity with high longevity deceased donor kidney transplantation under the {US Kidney Allocation System}.
\newblock \emph{American Journal of Kidney Diseases}, 84\penalty0 (4):\penalty0 416--426, 2024.

\bibitem[Ashby et~al.(2017)Ashby, Leichtman, Rees, Song, Bray, Wang, and Kalbfleisch]{ashby2017kidney}
Valarie~B. Ashby, Alan~B. Leichtman, Michael~A. Rees, Peter X.-K. Song, Mathieu Bray, Wen Wang, and John~D. Kalbfleisch.
\newblock A kidney graft survival calculator that accounts for mismatches in age, sex, {HLA}, and body size.
\newblock \emph{Clinical Journal of the American Society of Nephrology}, 12\penalty0 (7):\penalty0 1148--1160, 2017.

\bibitem[Bekbolsynov et~al.(2022)Bekbolsynov, Mierzejewska, Khuder, Ekwenna, Rees, {Green II}, and Stepkowski]{Bekbolsynov2022HLAImmunogenicity}
Dulat Bekbolsynov, Beata Mierzejewska, Sadik Khuder, Obinna Ekwenna, Michael Rees, Robert~C. {Green II}, and Stanislaw~M. Stepkowski.
\newblock Improving access to {HLA}-matched kidney transplants for {African American} patients.
\newblock \emph{Frontiers in Immunology}, 13:\penalty0 832488, 2022.
\newblock \doi{10.3389/fimmu.2022.832488}.
\newblock URL \url{https://doi.org/10.3389/fimmu.2022.832488}.

\bibitem[Clayton et~al.(2014)Clayton, McDonald, Snyder, Salkowski, and Chadban]{clayton2014external}
P.~A. Clayton, S.~P. McDonald, J.~J. Snyder, N.~Salkowski, and S.~J. Chadban.
\newblock External validation of the estimated posttransplant survival score for allocation of deceased donor kidneys in the {United States}.
\newblock \emph{American Journal of Transplantation}, 14\penalty0 (8):\penalty0 1922--1926, 2014.

\bibitem[Cox(1972)]{cox1972regression}
David~R. Cox.
\newblock Regression models and life-tables.
\newblock \emph{Journal of the Royal Statistical Society: Series B (Methodological)}, 34\penalty0 (2):\penalty0 187--202, 1972.

\bibitem[Cremers et~al.(2026)Cremers, Stewart, Massie, Segev, Gentry, and Mankowski]{cremers2026global}
Roby Cremers, Darren Stewart, Allan~B. Massie, Dorry~L. Segev, Sommer~E. Gentry, and Michal~A. Mankowski.
\newblock A global review of organ allocation simulation models.
\newblock \emph{Transplantation}, 110\penalty0 (3):\penalty0 e573--e582, 2026.

\bibitem[Fan et~al.(2010)Fan, Ashby, Fuller, Boulware, Kao, Norman, Randall, Young, Kalbfleisch, and Leichtman]{fan2010access}
P.-Y. Fan, Valarie~B. Ashby, D.~S. Fuller, L.~E. Boulware, A.~Kao, Silas~P. Norman, H.~B. Randall, C.~Young, John~D. Kalbfleisch, and Alan~B. Leichtman.
\newblock Access and outcomes among minority transplant patients, 1999--2008, with a focus on determinants of kidney graft survival.
\newblock \emph{American Journal of Transplantation}, 10\penalty0 (4p2):\penalty0 1090--1107, 2010.

\bibitem[Fotso(2018)]{fotso2018deepneuralnetworkssurvival}
Stephane Fotso.
\newblock Deep neural networks for survival analysis based on a multi-task framework.
\newblock \emph{arXiv preprint arXiv:1801.05512}, 2018.
\newblock URL \url{https://arxiv.org/abs/1801.05512}.

\bibitem[Gordon et~al.(2010)Gordon, Ladner, Caicedo, and Franklin]{gordon2010disparities}
Elisa~J. Gordon, Daniela~P. Ladner, Juan~Carlos Caicedo, and John Franklin.
\newblock Disparities in kidney transplant outcomes: a review.
\newblock \emph{Seminars in Nephrology}, 30\penalty0 (1):\penalty0 81--89, 2010.

\bibitem[Haider et~al.(2020)Haider, Hoehn, Davis, and Greiner]{haider2020effective}
Humza Haider, Bret Hoehn, Sarah Davis, and Russell Greiner.
\newblock Effective ways to build and evaluate individual survival distributions.
\newblock \emph{Journal of Machine Learning Research}, 21\penalty0 (85):\penalty0 1--63, 2020.

\bibitem[Harrell et~al.(1982)Harrell, Califf, Pryor, Lee, and Rosati]{harrell1982evaluating}
Frank~E. Harrell, Robert~M. Califf, David~B. Pryor, Kerry~L. Lee, and Robert~A. Rosati.
\newblock Evaluating the yield of medical tests.
\newblock \emph{JAMA}, 247\penalty0 (18):\penalty0 2543--2546, 1982.

\bibitem[Ishwaran et~al.(2008)Ishwaran, Kogalur, Blackstone, and Lauer]{ishwaran2008random}
Hemant Ishwaran, Udaya~B. Kogalur, Eugene~H. Blackstone, and Michael~S. Lauer.
\newblock Random survival forests.
\newblock \emph{Annals of Applied Statistics}, 2\penalty0 (3):\penalty0 841--860, 2008.

\bibitem[Kadatz et~al.(2023)Kadatz, Gill, Gill, Lan, McMichael, Chang, and Gill]{kadatz2023benefits}
Matthew~J. Kadatz, Jagbir Gill, Justin Gill, James~H. Lan, Lachlan~C. McMichael, Doris~T. Chang, and John~S. Gill.
\newblock The benefits of preemptive transplantation using high--{Kidney Donor Profile Index} kidneys.
\newblock \emph{Clinical Journal of the American Society of Nephrology}, 18\penalty0 (5):\penalty0 634--643, 2023.

\bibitem[Katzman et~al.(2018)Katzman, Shaham, Cloninger, Bates, Jiang, and Kluger]{katzman2018deepsurv}
Jared~L. Katzman, Uri Shaham, Alexander Cloninger, Jonathan Bates, Tingting Jiang, and Yuval Kluger.
\newblock {DeepSurv}: personalized treatment recommender system using a {Cox} proportional hazards deep neural network.
\newblock \emph{BMC Medical Research Methodology}, 18\penalty0 (1):\penalty0 24, 2018.
\newblock \doi{10.1186/s12874-018-0482-1}.

\bibitem[Lee et~al.(2018)Lee, Zame, Yoon, and van~der Schaar]{lee2018deephit}
Changhee Lee, William Zame, Jinsung Yoon, and Mihaela van~der Schaar.
\newblock {DeepHit}: A deep learning approach to survival analysis with competing risks.
\newblock In \emph{Proceedings of the AAAI Conference on Artificial Intelligence}, volume~32, 2018.
\newblock \doi{10.1609/aaai.v32i1.11842}.

\bibitem[Lee et~al.(2019)Lee, Kanellis, and Mulley]{lee2019allocation}
Darren Lee, John Kanellis, and William~R. Mulley.
\newblock Allocation of deceased donor kidneys: A review of international practices.
\newblock \emph{Nephrology}, 24\penalty0 (6):\penalty0 591--598, 2019.
\newblock \doi{10.1111/nep.13548}.

\bibitem[Lei et~al.(2024)Lei, Gohari, and Farnia]{lei2024inductivebiasesdemographicparitybased}
Haoyu Lei, Amin Gohari, and Farzan Farnia.
\newblock On the inductive biases of demographic parity-based fair learning algorithms.
\newblock In \emph{Proceedings of the 40th Conference on Uncertainty in Artificial Intelligence}, pages 2205--2225, 2024.
\newblock URL \url{https://proceedings.mlr.press/v244/lei24a.html}.

\bibitem[Lillelund et~al.(2026)Lillelund, Qi, Greiner, and Pedersen]{lillelund2025stop}
Christian~Marius Lillelund, Shi-ang Qi, Russell Greiner, and Christian~Fischer Pedersen.
\newblock Position: Stop chasing the {C-index} when evaluating survival analysis models.
\newblock \emph{arXiv preprint arXiv:2506.02075}, 2026.
\newblock URL \url{https://arxiv.org/abs/2506.02075}.

\bibitem[Mark et~al.(2019)Mark, Goldsman, Gurbaxani, Keskinocak, and Sokol]{mark2019kidney}
Ethan Mark, David Goldsman, Brian Gurbaxani, Pinar Keskinocak, and Joel Sokol.
\newblock Using machine learning and an ensemble of methods to predict kidney transplant survival.
\newblock \emph{PLOS ONE}, 14\penalty0 (1):\penalty0 e0209068, 2019.
\newblock \doi{10.1371/journal.pone.0209068}.
\newblock URL \url{https://doi.org/10.1371/journal.pone.0209068}.

\bibitem[Meng et~al.(2022)Meng, Wang, Zhang, Zhang, Zhang, Zhang, and Wang]{meng2022novel}
Xiangyu Meng, Xun Wang, Xudong Zhang, Chaogang Zhang, Zhiyuan Zhang, Kuijie Zhang, and Shudong Wang.
\newblock A novel attention-mechanism based {Cox} survival model by exploiting pan-cancer empirical genomic information.
\newblock \emph{Cells}, 11\penalty0 (9):\penalty0 1421, 2022.

\bibitem[Molinari et~al.(2022)Molinari, Kaltenmeier, Liu, Ashwat, Jorgensen, Puttarajappa, Wu, Mehta, Sood, Shah, Sharma, Thompson, Reddy, and Hariharan]{Molinari2022HighKDPI}
Michele Molinari, Christof Kaltenmeier, Hao Liu, Eishan Ashwat, Dana Jorgensen, Chethan Puttarajappa, Christine~M. Wu, Rajil Mehta, Puneet Sood, Nirav Shah, Akhil Sharma, Ann Thompson, Dheera Reddy, and Sundaram Hariharan.
\newblock Function and longevity of renal grafts from high-{KDPI} donors.
\newblock \emph{Clinical Transplantation}, 36\penalty0 (9):\penalty0 e14759, 2022.
\newblock \doi{10.1111/ctr.14759}.

\bibitem[Na et~al.(2025)Na, Koo, Koh, Kim, and Yang]{Na2025KDPIEPTS}
O.~Na, T.~Y. Koo, H.~B. Koh, B.~S. Kim, and J.~Yang.
\newblock Kidney transplant outcomes according to matching of the {Kidney Donor Profile Index} and {Estimated Post-Transplant Survival} scores.
\newblock \emph{Kidney Research and Clinical Practice}, 2025.
\newblock \doi{10.23876/j.krcp.25.083}.
\newblock URL \url{https://doi.org/10.23876/j.krcp.25.083}.

\bibitem[Nemati et~al.(2023)Nemati, Zhang, Sloma, Bekbolsynov, Wang, Stepkowski, and Xu]{nemati2023predicting}
Mohammadreza Nemati, Haonan Zhang, Michael Sloma, Dulat Bekbolsynov, Hong Wang, Stanislaw Stepkowski, and Kevin~S. Xu.
\newblock Predicting kidney transplant survival using multiple feature representations for {HLAs}.
\newblock \emph{Artificial Intelligence in Medicine}, 145:\penalty0 102675, 2023.

\bibitem[Opelz et~al.(1999)Opelz, Wujciak, D{\"o}hler, Scherer, and Mytilineos]{opelz1999hla}
Gerhard Opelz, Thomas Wujciak, Bernd D{\"o}hler, Sabine Scherer, and Joannis Mytilineos.
\newblock {HLA} compatibility and organ transplant survival. {Collaborative} transplant study.
\newblock \emph{Reviews in Immunogenetics}, 1\penalty0 (3):\penalty0 334--342, 1999.

\bibitem[{Organ Procurement and Transplantation Network}(2025)]{hrsa_unos_national_data}
{Organ Procurement and Transplantation Network}.
\newblock National data.
\newblock \url{https://hrsa.unos.org/data/view-data-reports/national-data/}, 2025.
\newblock Accessed: 2026-04-17.

\bibitem[Paquette et~al.(2022)Paquette, Ghassemi, Bukhtiyarova, Cisse, Gagnon, Della~Vecchia, Rabearivelo, and Loudiyi]{paquette2022kidney}
Fran{\c{c}}ois-Xavier Paquette, Amir Ghassemi, Olga Bukhtiyarova, Moustapha Cisse, Natanael Gagnon, Alexia Della~Vecchia, Hobivola~A. Rabearivelo, and Youssef Loudiyi.
\newblock Machine learning support for decision-making in kidney transplantation: Step-by-step development of a technological solution.
\newblock \emph{JMIR Medical Informatics}, 10\penalty0 (6):\penalty0 e34554, 2022.
\newblock \doi{10.2196/34554}.
\newblock URL \url{https://doi.org/10.2196/34554}.

\bibitem[Poggio et~al.(2021)Poggio, Augustine, Arrigain, Brennan, and Schold]{poggio2021longterm}
Emilio~D. Poggio, John~J. Augustine, S.~Arrigain, Daniel~C. Brennan, and Jesse~D. Schold.
\newblock Long-term kidney transplant graft survival—making progress when most needed.
\newblock \emph{American Journal of Transplantation}, 21\penalty0 (8):\penalty0 2824--2832, 2021.
\newblock \doi{10.1111/ajt.16463}.

\bibitem[Ponticelli(2015)]{ponticelli2015impact}
Claudio~E. Ponticelli.
\newblock The impact of cold ischemia time on renal transplant outcome.
\newblock \emph{Kidney International}, 87\penalty0 (2):\penalty0 272--275, 2015.

\bibitem[Simon et~al.(2011)Simon, Friedman, Hastie, and Tibshirani]{simon2011regularization}
Noah Simon, Jerome~H. Friedman, Trevor Hastie, and Rob Tibshirani.
\newblock Regularization paths for {Cox's} proportional hazards model via coordinate descent.
\newblock \emph{Journal of Statistical Software}, 39:\penalty0 1--13, 2011.

\bibitem[Truchot et~al.(2023)Truchot, Raynaud, Kamar, Naesens, Legendre, Delahousse, Thaunat, Buchler, Crespo, Linhares, Orandi, Akalin, Pujol, Silva~Jr, Gupta, Segev, Jouven, Bentall, Stegall, Lefaucheur, and Loupy]{Truchot2023MLvsStats}
A.~Truchot, M.~Raynaud, N.~Kamar, M.~Naesens, C.~Legendre, M.~Delahousse, O.~Thaunat, M.~Buchler, M.~Crespo, K.~Linhares, B.~J. Orandi, E.~Akalin, G.~S. Pujol, H.~T. Silva~Jr, G.~Gupta, D.~L. Segev, X.~Jouven, A.~J. Bentall, M.~D. Stegall, C.~Lefaucheur, and A.~Loupy.
\newblock Machine learning does not outperform traditional statistical modelling for kidney allograft failure prediction.
\newblock \emph{Kidney International}, 103\penalty0 (5):\penalty0 936--948, 2023.
\newblock \doi{10.1016/j.kint.2022.12.011}.
\newblock URL \url{https://doi.org/10.1016/j.kint.2022.12.011}.

\bibitem[van~de Klundert et~al.(2025)van~de Klundert, Perez-Galarce, Olivares, Pengel, and de~Weerd]{vandeKlundert2025comparative}
Jeroen van~de Klundert, Francisco Perez-Galarce, Mauricio Olivares, Lily Pengel, and Arie de~Weerd.
\newblock The comparative performance of models predicting patient and graft survival after kidney transplantation: A systematic review.
\newblock \emph{Transplantation Reviews}, 39\penalty0 (3):\penalty0 100934, 2025.
\newblock \doi{10.1016/j.trre.2025.100934}.

\bibitem[Wolfe et~al.(2009)Wolfe, McCullough, and Leichtman]{wolfe2009predictability}
R.~A. Wolfe, K.~P. McCullough, and A.~B. Leichtman.
\newblock Predictability of survival models for waiting list and transplant patients: Calculating {LYFT}.
\newblock \emph{American Journal of Transplantation}, 9:\penalty0 1523--1527, 2009.
\newblock \doi{10.1111/j.1600-6143.2009.02708.x}.
\newblock URL \url{https://doi.org/10.1111/j.1600-6143.2009.02708.x}.

\bibitem[Wolfe et~al.(2008)Wolfe, McCullough, Schaubel, Kalbfleisch, Murray, Stegall, and Leichtman]{wolfe2008lyft}
Robert~A. Wolfe, Keith~P. McCullough, Douglas~E. Schaubel, Jack~D. Kalbfleisch, Susan Murray, Mark~D. Stegall, and Alan~B. Leichtman.
\newblock Calculating life years from transplant ({LYFT}): Methods for kidney and kidney-pancreas candidates.
\newblock \emph{American Journal of Transplantation}, 8\penalty0 (4p2):\penalty0 997--1011, 2008.

\bibitem[Yu et~al.(2011)Yu, Greiner, Lin, and Baracos]{yu2011learning}
Chun-Nam Yu, Russell Greiner, Hsiu-Chin Lin, and Vickie Baracos.
\newblock Learning patient-specific cancer survival distributions as a sequence of dependent regressors.
\newblock In \emph{Advances in Neural Information Processing Systems}, volume~24, 2011.

\bibitem[Zens et~al.(2018)Zens, Danobeitia, Leverson, Chlebeck, Zitur, Redfield, D'Alessandro, Odorico, Kaufman, and Fernandez]{zens2018impact}
T.~J. Zens, J.~S. Danobeitia, G.~Leverson, P.~J. Chlebeck, L.~J. Zitur, R.~R. Redfield, A.~M. D'Alessandro, S.~Odorico, D.~B. Kaufman, and L.~A. Fernandez.
\newblock The impact of kidney donor profile index on delayed graft function and transplant outcomes: A single-center analysis.
\newblock \emph{Clinical Transplantation}, 32:\penalty0 e13190, 2018.
\newblock \doi{10.1111/ctr.13190}.

\bibitem[Zhang et~al.(2023)Zhang, Deng, Muller, Wong, and Yang]{zhang2023multi}
Yunwei Zhang, Danny Deng, Samuel Muller, Germaine Wong, and Jean Yee~Hwa Yang.
\newblock A multi-step precision pathway for predicting allograft survival in heterogeneous cohorts of kidney transplant recipients.
\newblock \emph{Transplant International}, Volume, 2023.
\newblock \doi{10.3389/ti.2023.11338}.
\newblock URL \url{https://www.frontierspartnerships.org/journals/transplant-international/articles/10.3389/ti.2023.11338}.

\bibitem[Zhou et~al.(2023)Zhou, Wang, Wang, and Zou]{zhou2023survmetrics}
Hanpu Zhou, Hong Wang, Sizheng Wang, and Yi~Zou.
\newblock {SurvMetrics}: An {R} package for predictive evaluation metrics in survival analysis.
\newblock \emph{R Journal}, 14\penalty0 (4):\penalty0 252--263, 2023.

\end{thebibliography}

\newpage
\appendix

\section{Hyperparameter Tuning for ML-based Predictors}
\label{sec:hp_tuning}
\appendix
For all models, we perform a grid search over the predefined hyperparameter spaces listed in Table \ref{tab:hp_grid}. Model selection is based on validation performance using the nested cross-validation scheme described in Section \ref{sec:ml_predictors}. Specifically, we employ 5-fold outer cross-validation for model evaluation, with an inner 2-fold cross-validation loop for hyperparameter tuning. 
Early stopping is applied where applicable to prevent overfitting, using the inner validation splits to monitor performance during training. The best-performing configuration from the inner CV is selected and evaluated on the corresponding outer CV fold.

\begin{table}[t]
\centering
\caption{Summary of hyperparameter grids for all models.}
\label{tab:hp_grid}
\begin{tabular}{lll}
\toprule
\textbf{Model} & \textbf{Hyperparameter} & \textbf{Values} \\
\midrule
CoxNet & $l_1$ ratio & $\{0.1, 0.3, 0.5\}$ \\
        & $\alpha$ & $\{0.01, 0.05, 0.1\}$ \\

DeepHit & Hidden dim & $\{32, 64\}$ \\
        & \# bins & $\{5, 10, 15, 20\}$ \\
        & $\alpha$ & $\{0.1, 0.5, 1.0\}$ \\
        & $\sigma$ & $\{0.1, 0.5, 1.0\}$ \\
        & LR & $\{0.01\}$ \\

DeepSurv & Architecture & $\{[16], [32], [16,16], [32,32]\}$ \\
         & Dropout & $\{0.1, 0.3, 0.5\}$ \\
         & Batch norm & True \\
         & Optimizer & Adam \\
         & LR & $\{0.01\}$ \\
         & Batch size & $256$ \\

N-MTLR & Architecture & $\{[16], [32], [16,16], [32,32]\}$ \\
       & Dropout & $\{0.1, 0.3, 0.5\}$ \\
       & Batch norm & True \\
       & Optimizer & Adam \\
       & LR & $\{0.01\}$ \\
       & Batch size & $256$ \\
       & \# bins & $10$ \\

RSF & $n_{\text{estimators}}$ & $\{100, 200, 300\}$ \\
    & Max depth & $\{5,10,15,20,25,30\}$ \\
    & Max features & $\sqrt{\cdot}$ \\
\bottomrule
\end{tabular}
\end{table}

\end{document}